\documentclass[letterpaper, 10 pt, conference]{ieeeconf}  

\usepackage[absolute, overlay]{textpos}
\textblockorigin{0mm}{3mm}

\usepackage{cite}
\usepackage[dvipdfmx]{graphicx}
\usepackage{here}
\usepackage{physics}
\usepackage{hhline}
\usepackage[dvipdfmx]{color}

\IEEEoverridecommandlockouts                              

\title{\LARGE \bf
Object Pose Estimation by Camera Arm Control 
Based on the Next Viewpoint Estimation
}

\author{Tomoki Mizuno$^{1}$, Kazuya Yabashi$^{2}$, Tsuyoshi Tasaki$^{3}$%
\thanks{*This work was not supported by any organization}%
\thanks{$^{1}$Meijo University ,1-501 Shiogamaguchi,Tenpaku-ku, Nagoya, Aichi
        {\tt\small 200442160@ccmailg.meijo-u.ac.jp}}%
\thanks{$^{2}$Meijo University ,1-501 Shiogamaguchi,Tenpaku-ku, Nagoya, Aichi
        {\tt\small 190442146@ccmailg.meijo-u.ac.jp}}%
\thanks{$^{3}$Meijo University ,1-501 Shiogamaguchi,Tenpaku-ku, Nagoya, Aichi
        {\tt\small tasaki@meijo-u.ac.jp}}%
}

\begin{document}

\maketitle
\thispagestyle{empty}
\pagestyle{empty}

\begin{abstract}

We have developed a new method to estimate a Next Viewpoint (NV) which is effective for pose estimation of simple-shaped products for product display robots in retail stores. Pose estimation methods using Neural Networks (NN) based on an RGBD camera are highly accurate, but their accuracy significantly decreases when the camera acquires few texture and shape features at a current view point. However, it is difficult for previous mathematical model-based methods to estimate effective NV which is because the simple shaped objects have few shape features. Therefore, we focus on the relationship between the pose estimation and NV estimation. When the pose estimation is more accurate, the NV estimation is more accurate. Therefore, we develop a new pose estimation NN that estimates NV simultaneously. Experimental results showed that our NV estimation realized a pose estimation success rate 77.3\%, which was 7.4pt higher than the mathematical model-based NV calculation did. Moreover, we verified that the robot using our method displayed 84.2\% of products.\\

\end{abstract}


\begin{textblock}{19}(1,0)
        \copyright 2024 IEEE.  Personal use of this material is permitted.  Permission from IEEE must be obtained for all other uses, in any current or future media, including reprinting/republishing this material for advertising or promotional purposes, creating new collective works, for resale or redistribution to servers or lists, or reuse of any copyrighted component of this work in other works.

        Published in: 2024 IEEE/RSJ International Conference on Intelligent Robots and Systems

        DOI: 10.1109/IROS58592.2024.10801633
\end{textblock}

\section{INTRODUCTION}

Currently, the labor shortage in retail stores is becoming severe, and there is a growing expectation for the automation of product display using robots\cite{c1,c2,c3}. Automating product display requires the estimation of object poses, typically using color and depth images acquired with RGBD cameras\cite{c4}\hspace{-1pt}\cite{c5}. Recently, methods using Neural Networks (NNs) with color and depth images as inputs have been developed to estimate poses with high accuracy\cite{pynet,dense_fusion,ffb6d}. Among them, PYNet\cite{pynet} accurately estimates the poses of simple shaped objects commonly found in retail stores, such as rectangular prisms, cylinders and triangular prisms. However, general pose estimation methods decrease accuracy depending on the position of the camera relative to the object, especially for unknown objects not included in the training data. 
The pose estimation accuracy decreases when there is less information available from the camera due to the product's pose. Especially for simple-shaped objects, it is difficult to estimate the pose when only images like Fig. 1 are obtained. Therefore, changing the position of the camera mounted on the robot improves the pose estimation accuracy.\\

In SLAM (Simultaneously Localization And Mapping) and SfM (Structure from Motion), there are many studies on determining the next viewpoint to increase the amount of information obtained from sensors\cite{long_edge}\hspace{-1pt}\cite{outline}. For example, Simon Kriegel et al. increased the amount of information by searching for areas where long edges could be obtained in images, efficiently conducting SfM\cite{long_edge} . However, as shown in Fig. 1, it is often impossible to uniquely determine the direction in which information increases for simple-shaped objects.
Therefore, this study tackles the novel challenge of estimating an effective next viewpoint for pose estimation of simple-shaped objects with the goal of automating product display by robots.
To address the challenge, this study focuses on the relationship between object pose and effective next viewpoints. If the pose can be correctly estimated, it is possible to correctly estimate the effective viewpoint. Conversely, the direction of the effective viewpoint can also work as a guideline for pose estimation. This study develops a new method that allows the pose estimation NN to simultaneously estimate a next viewpoint, realizing the automation of product display.\\

The academic contributions of this study are as follows:

\begin{figure}[t]
\begin{center}
\includegraphics[width=0.8\linewidth]{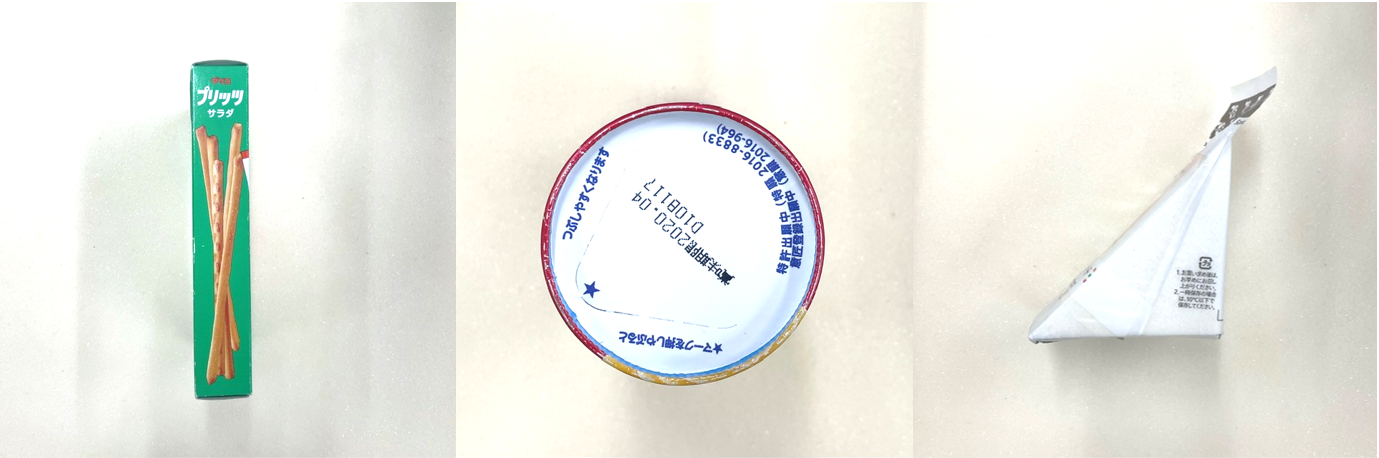}
\caption{difficult example of pose estimation}
\end{center}
\end{figure}   

\begin{itemize}
\item Developed a new NN that estimates a next viewpoint simultaneously with the pose estimation.
\item Demonstrated that pose estimation accuracy improves by simultaneously estimating a next viewpoint.
\item Showed that the next viewpoint estimation by the NN considering the pose is effective compared to the next viewpoint estimation based on the mathematical model. 
\item Implemented the developed next viewpoint estimation method into a robot, demonstrated improving the number of displaying products successfully. \\
\end{itemize}

\section{RELATED RESEARCH}

\subsection{RGBD Camera Pose Estimation}

With the release of the LINEMOD dataset\cite{linemod}, many NNs for the pose estimation using RGBD cameras have been developed, such as DeepIM\cite{deep_im}, DenseFusion\cite{dense_fusion} and FFB6D\cite{ffb6d}. Methods with the high accuracy in the LINEMOD dataset \cite{linemod} estimate the poses of complex-shaped objects accurately. However, for simple-shaped objects with few shape features, as shown in  Fig. 1, the accuracy decreases.\\\par
PYNet\cite{pynet} has been more accurate than FFB6D\cite{ffb6d}, which once achieved the highest accuracy in the LINEMOD dataset, in the pose estimation of simple-shaped objects. The representative shapes of simple-shaped objects are rectangular prisms, cylinders and triangular prisms as shown in Fig. 2. PYNet\cite{pynet} estimates the surface on which the object is grounded and the angle around the axis perpendicular to the grounding surface. PYNet\cite{pynet} calls the surface “poseclass” and the angle “yaw angle”. By estimating the poseclass, the 3D pose estimation problem is simplified to a 1D yaw angle estimation problem, which improves the pose estimation accuracy. The poseclass is estimated by solving a classification problem with NNs, dividing rectangular prisms, cylinders, and triangular prisms into 6, 8 and 5 classes, respectively, as shown in Fig. 2. PYNet\cite{pynet} divides the sides of cylinders into 6 classes because it thinks 60deg is the necessary resolution for product display by a robot. The left side of Fig. 2 shows the net of each simple shape product and the assigned poseclasses to the unfolded product surfaces. The right side of Fig. 2 shows images of the products taken from above when the assigned poseclasses are grounded. By estimating the poseclass and outputting the yaw angle as shown in Fig. 3, the 3D pose is determined. In this study, we refer to  PYNet\cite{pynet} architecture for the purpose of display simple-shaped objects by robots.\\

\subsection{Next Viewpoint Estimation}

The next viewpoint estimation has been extensively researched in SLAM and SfM. A simple edge-based method uses the direction of long edges, where information is abundant\cite{long_edge}. The edge-based method\cite{long_edge} detects the longest edge observable from the current viewpoint and searches for the next viewpoint in the direction of the longest edge. However, the simple-shaped objects often have multiple long edge. As shown in Fig. 4, moving the camera in the direction of the product's front may increase accuracy, but there are cases where moving to the backside. Therefore, even if the viewpoint is moved in the direction of long edges, the accuracy may not always improve in pose estimation.\\\par
In SfM, there is an outlier-based method\cite{outline} that moves the viewpoint to areas with many outliers in the 3D point cloud. The outlier-based method is utilized for 3D modeling objects in SfM, because the area with many outliers is the complex shape area and requires many observations. However, it is difficult to apply the outlier-based method to the pose estimation of simple-shaped objects because simple-shaped objects have fewer outliers.
        
\begin{figure}[t]
\begin{center}
\includegraphics[width=0.95\linewidth]{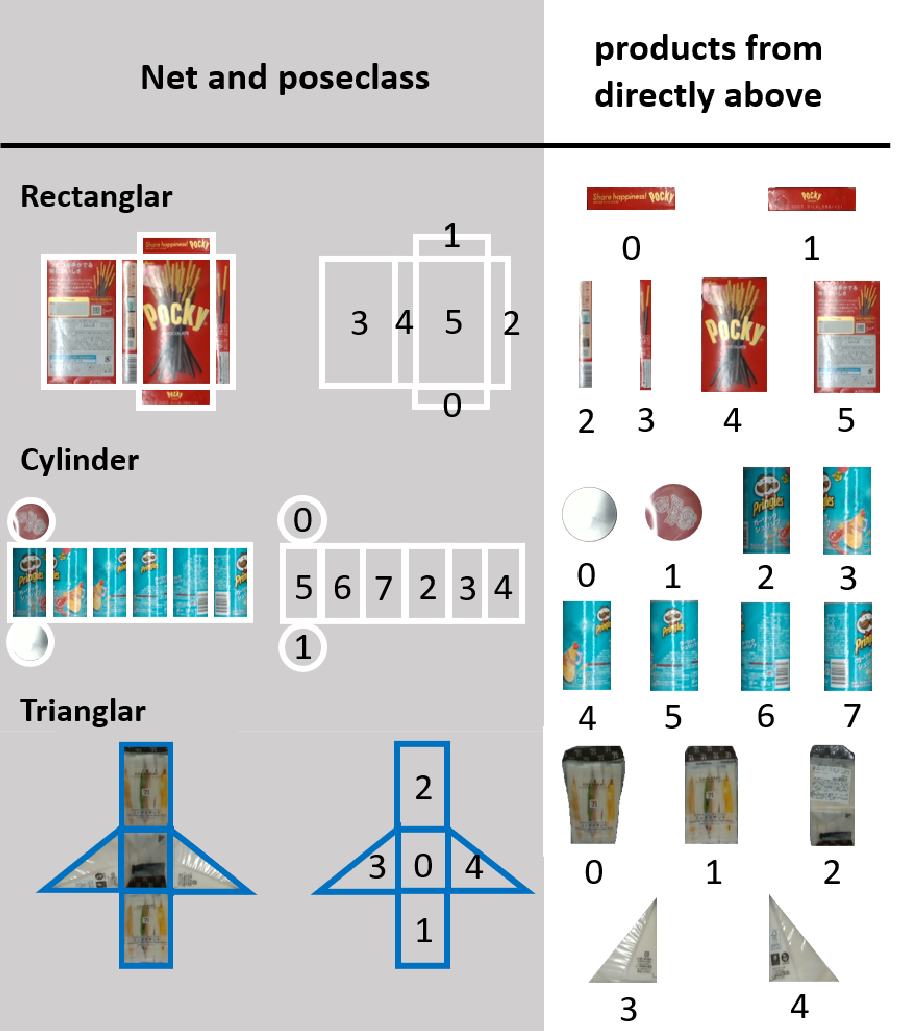}
\caption{poseclass of the simple-shaped product}
\end{center}
\end{figure}

\begin{figure}[t!]
\begin{center}

\includegraphics[width=0.8\linewidth]{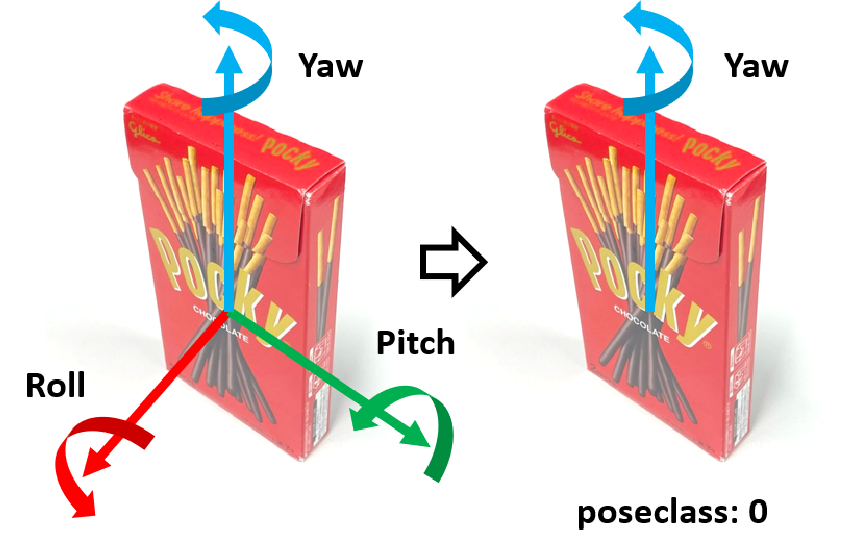}
\caption{pose estimation using poseclass}
\includegraphics[width=0.8\linewidth]{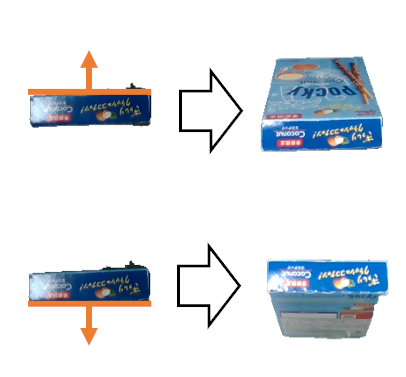}
\caption{example of camera motion}
        
\end{center}
\end{figure}

\begin{figure*}[t]
\begin{center}
\includegraphics[width=1.0\linewidth]{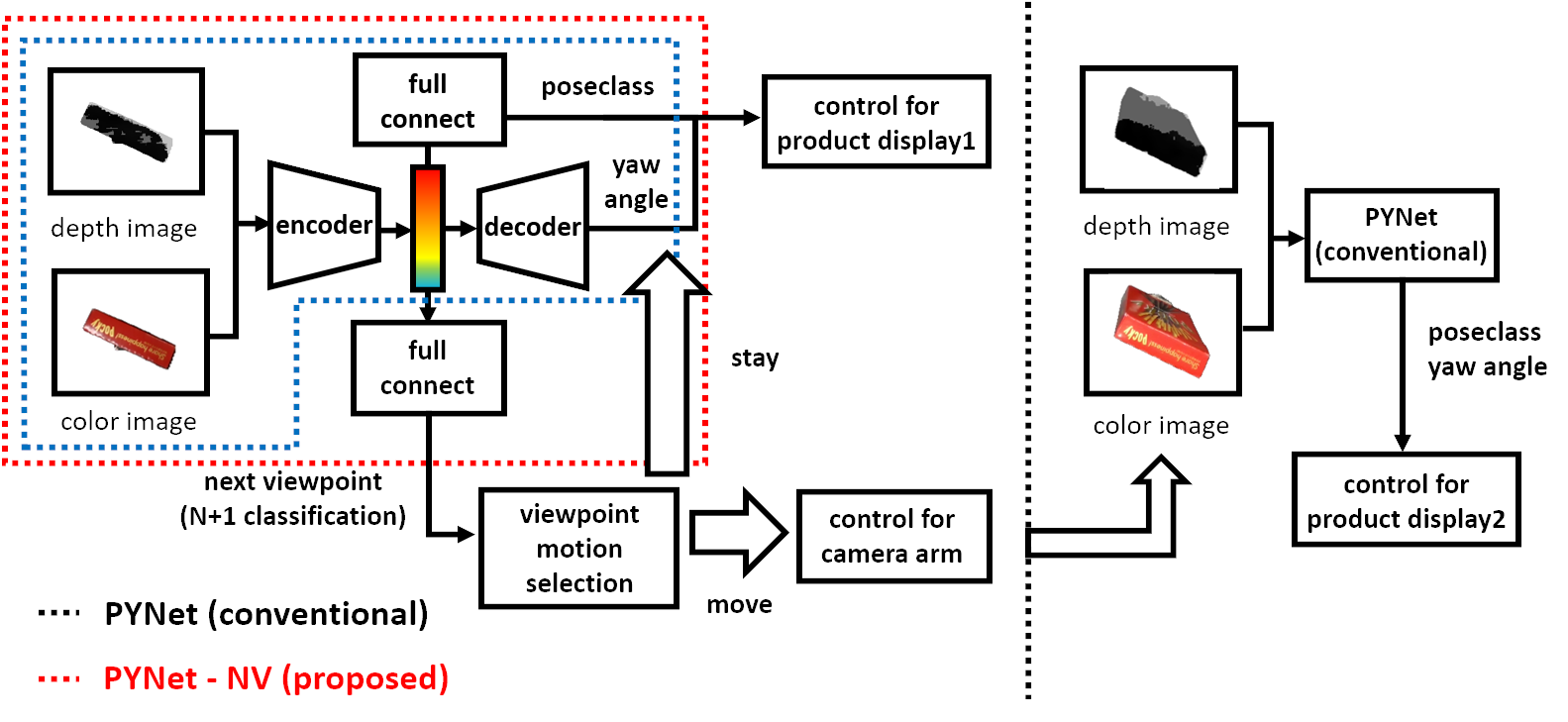}
\caption{system using PYNet estimating Next Viewpoint (PYNet-NV)}
\end{center}
\end{figure*}

\begin{figure}[t]
\begin{center}
\includegraphics[width=0.8\linewidth]{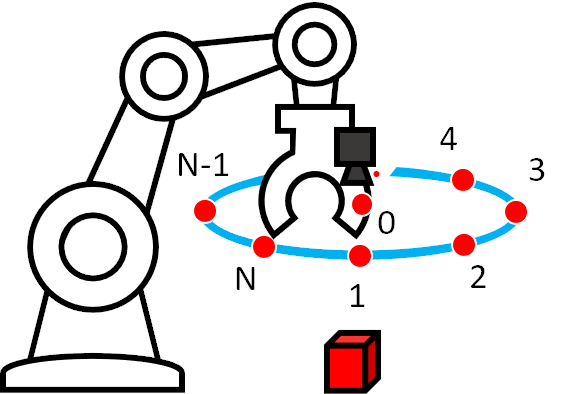}
\caption{next viewpoint candidates}
\end{center}
\end{figure}

\subsection{Positioning this study}

There are many studies on the pose estimation and next viewpoint estimation \cite{pynet,dense_fusion,ffb6d,long_edge,outline}. However, there has been no research on the next viewpoint estimation to improve the pose estimation accuracy of simple-shaped objects.
This study develops a new NN that simultaneously estimates the pose and next viewpoint.

\section{Proposed Method}

The system structure that uses our proposed method is shown in Fig. 5. In this study, the viewpoint is estimated by the pose estimation NN to control the camera arm. This chapter explains the proposed NN and the training method for viewpoints in Section I\hspace{-1.2pt}I\hspace{-1.2pt}I-A and I\hspace{-1.2pt}I\hspace{-1.2pt}I-B, respectively.

\subsection{PYNet-NV and Arm Control} 

The new method making PYNet\cite{pynet} estimate the viewpoint is called PYNet-NV (PYNet\cite{pynet} estimating Next Viewpoint). The architecture of PYNet-NV and the structure of the developed system are shown in Fig. 5. In this study, we refer to the PYNet's architecture for the pose estimation of simple-shaped objects because of its high accuracy. In our system, PYNet-NV estimates the effective viewpoint for the pose estimation, and after moving to the estimated viewpoint, the second pose estimation is performed. The conventional PYNet\cite{pynet} is used for the second pose estimation. When the viewpoint that the robot doesn't have to move is estimated, the first estimated pose is used.
To shorten the operation time by the robot, the viewpoint movement is limited to once. The robot controls the arm based on the results of the first or second pose estimation, respectively, in "control for product display 1 or 2" in Fig. 5, which performs product display.
In this study, the initial viewpoint for the pose estimation is directly above the object which is detected by object detection methods\cite{rembg}\hspace{-1pt}\cite{sam}. If candidates of next viewpoints involve positions that require complex three-dimensional movement, only robots with high degrees of freedom can work. Therefore, the next viewpoint candidates are $N$ viewpoints which set evenly on the circle with radius $r$ that can be moved flatly as shown in Fig. 6. Note that, the candidate also includes the initial viewpoint directly above the object, as shown in Fig. 6 “0”. Therefore, the next viewpoint estimation can be solved as an $N+1$ classification problem.
 
PYNet-NV leverages features effective for the pose estimation by branching from the encoder layer of PYNet\cite{pynet}. The branched viewpoint estimation part is realized with a single fully connected layer.\\

\subsection{Training Method for Viewpoint Estimation Branch}

PYNet-NV is trained to select a viewpoint effective for the pose estimation. In the case of simple shapes, the information effective for pose estimation is the texture. Therefore, PYNet-NV trained to select a viewpoint where the robot can see surfaces with complex textures, namely those with many edges in the images.
Teaching data for the next viewpoint is created based on the edge amount. Specifically, an $N+1$ dimensional one-hot vector, where only the element $v$ becomes 1, is used as the teaching data for the next viewpoint estimation, as shown in (\ref{move}).

\begin{equation}
        \label{move}
        v= argmax_{i \in{}V}e_i    
\end{equation}

Here, $e_{i}$ denotes the number of pixels indicating edges in the image obtained at the next viewpoint $i$, and $V$ denotes the set of next viewpoints. Note that, for each shape, the direct-above viewpoint is selected regardless of the edge amount when the difference of an object area in the image between before and after moving is small. Specifically, when the object grounding surface is poseclass $p$ that satisfies (\ref{non-move}), a one-hot vector with $v=0$ is used as the teaching data.

\begin{equation}
        \label{non-move}
        p= argmin_{k \in{}P}(max_{i \in{}V}a_{i}^{(k)}/a_{0}^{(k)})    
\end{equation}

Here, $P$ denotes the set of poseclass, and $a_{i}^{(k)}$ denotes the area of the detection rectangle of the object that can be observed from a viewpoint $i$ when the object is grounded in poseclass $k$.
An example of the teaching data for $N=4$ is shown in Fig. 7. It can be seen that the image obtained from viewpoints where the front of the product has many edges, which is more effective for the pose estimation than the direct-above viewpoint.
Since a next viewpoint is estimated as a classification problem, the cross-entropy $L_{v}$ is used as the loss function, as shown in (\ref{loss}).

\begin{equation}
        \label{loss}
        L_{v} = -\sum\limits_{j} \vb{t}(j)\log \vb{z}(j)
\end{equation}

In (3), $\vb{t}$ denotes the teaching data, and $\vb{z}$ denotes the vector indicating the estimated next viewpoint.\\

\begin{figure}[t!]
\begin{center}
\includegraphics[width=0.95\linewidth]{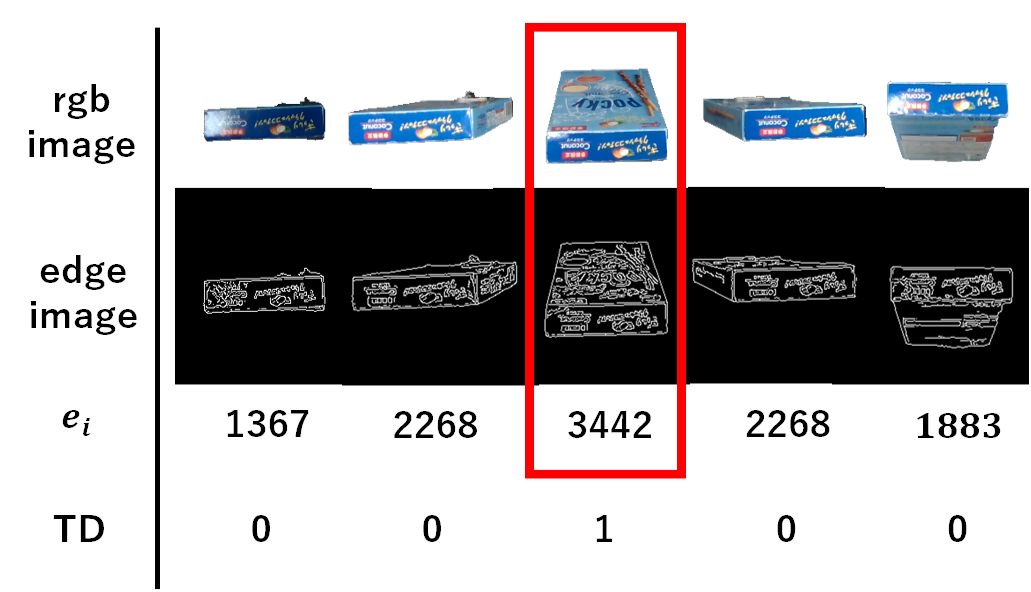}
\caption{example teaching data (TD)}
\vspace{10mm}
\includegraphics[width=0.9\linewidth]{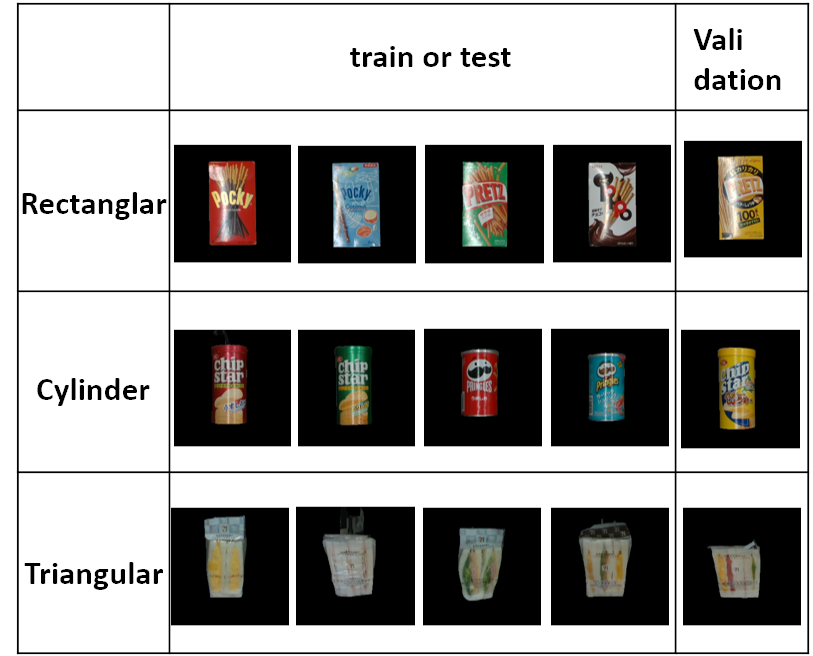}
\caption{products used in the experiment}

\end{center}
\end{figure}

\section{Evaluation on Pose Estimation Accuracy}

\subsection{Experimental Setup}

To confirm the effect of the viewpoint estimation with the pose estimation, we compare following 3 methods.\\
-PYNet-NV\\
-the edge-based method\cite{long_edge}\\
-PYNet\cite{pynet} \\
PYNet-NV and the edge-based method change a view point maximum once, and candidates of viewpoints of both methods are same. PYNet\cite{pynet} estimates a pose from only direct-above viewpoint.
The products used for training by PYNet-NV and PYNet\cite{pynet} are shown in Fig. 8. Similar to PYNet\cite{pynet}'s evaluation, this study uses rectangular prisms, cylinders and triangular prisms. The test is performed by leave-one-out cross-validation. That is, three types of products shown in the train or test column of Fig. 8 are used for training, and the remaining one type is used for test. Therefore, the test products are unknown products. The products shown in the validation column of Fig. 8 are used for validation during training. The NN parameters with the highest pose estimation accuracy on the validation data are used for test.
The data used for training, test and validation is obtained by capturing the RGBD images of the products with an RGBD camera (RealSense) from each viewpoint. During capturing, each product is rotated from 0 deg to 359 deg in 1-degree increments for each poseclass. As a result, we can obtain 360 sets of color and depth images of each product for each poseclass. The 360 sets of images are obtained for all viewpoints. The obtained image is cropped to a size of $224\times224$ by using object detection\cite{rembg}\hspace{-1pt}\cite{sam}. Therefore, the center of the image corresponds to the center of the product, approximately. 
Considering the manipulation range of the robot used in Chapter 5, the number of viewpoints $N$ and the radius $r$ are set to 4 and 0.15m, respectively. The evaluation index uses the angle error $e$ of the 3D pose. Using the angle error $e$, the proportion of outputs that satisfies (\ref{angle}) for all test data was evaluated as the success rate. $\phi$ is the arbitrarily set as the permissible angle error.

\begin{equation}
        \label{angle}
        e <= \phi
\end{equation}

In this study, $\phi$ is set from 0 deg to 30 deg in 0.1-degree increments to calculate the success rate. \\

\subsection{Experimental Results and Discussion}

The success rates of PYNet\cite{pynet}, the edge-based method\cite{long_edge} and PYNet-NV are shown in Fig. 9. Following the World Robot Summit\cite{wrs}, the discussion is based on the accuracy rate $\phi$ of 30 deg. As shown in Fig. 9, the success rates of PYNet\cite{pynet}, edge-based method and PYNet-NV were 68.0\%, 69.9\%, and 77.3\%, respectively. The pose estimation accuracy improved by changing viewpoint because the accuracy of edge-based method and PYNet-NV was higher than that of PYNet\cite{pynet}. Moreover, PYNet-NV improved accuracy by 7.4 points compared to the edge-based method. Therefore, it was found that estimating the next viewpoint by NN improved the pose estimation accuracy. Especially, the accuracy was higher when poseclass 0 and 1 of rectangular prisms and cylinders were grounded, which had low features of the shape.
Furthermore, the success rates of PYNet-NV and PYNet\cite{pynet} are shown in Fig. 10 when the evaluation was performed only with the test data for poseclass satisfying (2). In this case, PYNet-NV does not change the viewpoint and estimates pose just once. The poseclass satisfying (2) represents surfaces with little shape change even if the viewpoint change, such as the front of the object, where relatively large surfaces are visible. As shown in Fig. 10, even without changing a viewpoint, the success rate of PYNet-NV was 82.3\%, improving by 3.5 points over PYNet\cite{pynet}. Therefore, it is considered that there is an effect improving the pose estimation accuracy by estimating the next viewpoint and pose simultaneously.

\section{Evaluation of product display by robots}

\subsection{Experimental Setup}


To examine the impact of the improved pose estimation accuracy on the product display robot, we compare the robot using PYNet-NV with the robot using PYNet\cite{pynet}. The training setup for PYNet-NV and PYNet\cite{pynet} is the same as in Chapter 4.
The products used in this evaluation are those in the first column of each shape shown in Fig. 7. A yaw angle is determined randomly, and evaluations are conducted twice for each poseclass. The process of displaying the product to the shelf by the robot is shown in Fig. 11. In Fig. 11(a), the first pose estimation and next viewpoint estimation are performed. Then, in Fig. 11(b), the robot moves to the estimated next viewpoint and performs the second pose estimation. From Fig. 11(c) to (d), the product is displayed to the shelf so that the front of the product is visible based on the second pose estimation result. The evaluation is based on whether the product is successfully displayed facing the front on the shelf, as shown in Fig. 12.

\begin{figure}[t!]
\begin{center}
\includegraphics[width=0.8\linewidth]{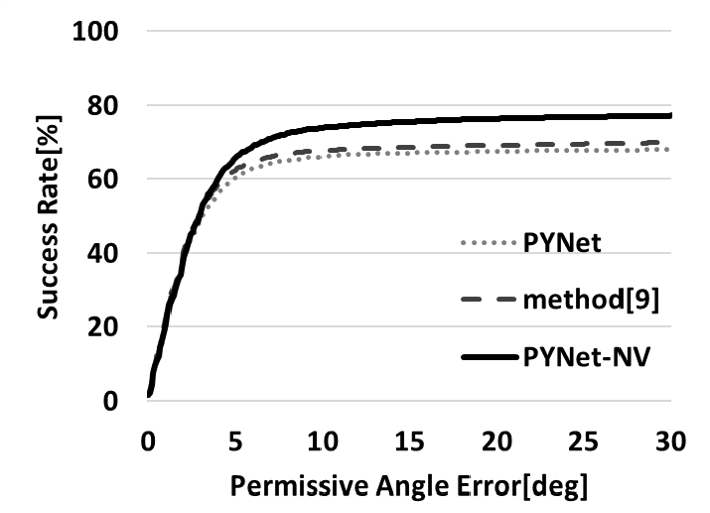}
\caption{success rate}
\includegraphics[width=0.8\linewidth]{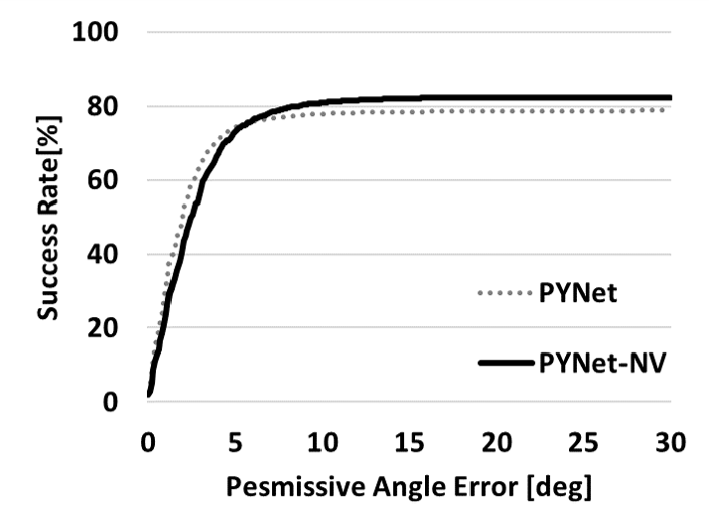}
\caption{success rate in non-moving poseclass}
\end{center}
\end{figure}

\begin{figure}[t!]
\begin{center}
\includegraphics[width=0.8\linewidth]{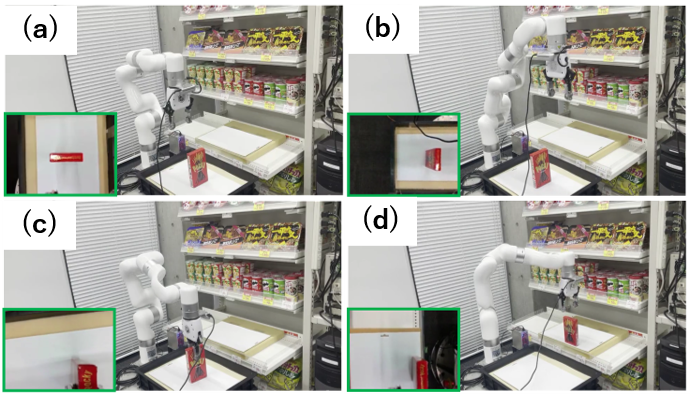}
\caption{product display flow}
\vspace{5mm}
\includegraphics[width=0.8\linewidth]{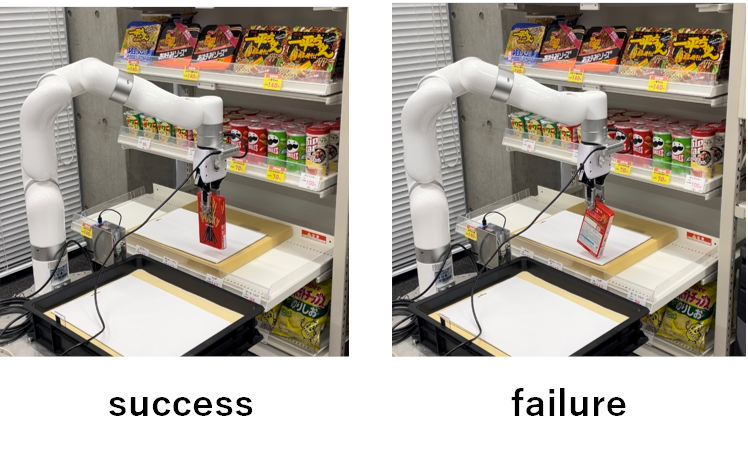}
\caption{example of success and failure}
\end{center}
\end{figure}

\subsection{Products used in the experiment}

The display results of PYNet-NV and PYNet\cite{pynet} are shown in Table.1. The product display was successful in 29 out of 38 times (76.3\%) by using PYNet\cite{pynet}. The product display was successful in 32 out of 38 times (84.2\%) by using PYNet-NV. Therefore, changing viewpoint based on the viewpoint estimation result is considered to contribute to the improvement of displaying performance.\par
Next, we want to discuss the number of success for each poseclass of the product. The number of successful product display for poseclass 1 of all products (Fig. 13) increased by using PYNet-NV compared to by using PYNet\cite{pynet}. Since poseclass 1 is a surface with little information, the changing viewpoint is considered to contribute to the product display performance.
However, as shown in Fig. 14, when poseclass 3 of the rectangular prism was grounded, the number of successful product display was zero. As shown in Fig. 14, products grounded in poseclass 3 have significantly different designs depending on the products, such as the presence or absence of barcodes. Even if an effective surface for the pose estimation becomes visible due to changing a viewpoint, pose estimation errors occur. This is because the most visible surface close to the camera includes distinctive designs and affects the pose estimation. Therefore, in the future, we plan to develop a pose estimation NN that focuses on the newly visible surfaces after changing a viewpoint.

\section{CONCLUSIONS}

\setcounter{figure}{12}
\begin{figure}[t!]
\begin{center}
\includegraphics[width=0.9\linewidth]{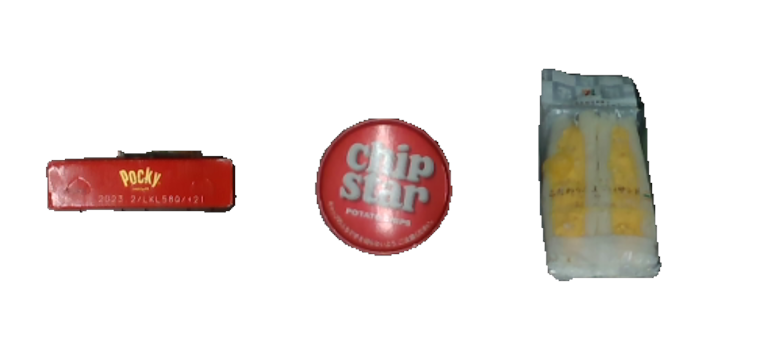}
\caption{poseclass 1 of all products}
\vspace{10mm}
\includegraphics[width=0.9\linewidth]{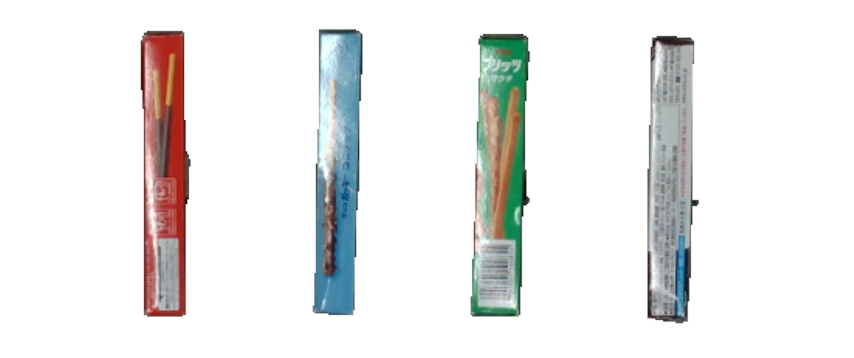}
\caption{poseclass 3 of rectanglar}
\end{center}
\end{figure}

\begin{table}[t!]
\begin{center}
\caption{results of displaying products}
\begin{tabular}{c||c|c|c|c} \hline
        \quad & Rectangular & Cylinder & Trianglar & Total \\ \hline\hline
        PYNet & 66.7\% & 87.5\% & 70.0\% & 76.3\% \\ 
        PYNet-NV & 83.3\% & 87.5\% & 80.0\% & 84.2\% \\ \hline
\end{tabular}
\end{center}
\end{table}

This study tackled the challenge of estimating an effective viewpoint for the pose estimation of simple-shaped objects with the goal of developing product display robots. We have developed PYNet-NV that estimates the product pose and next viewpoint simultaneously. The accuracy of the pose estimation based on a viewpoint estimated by using PYNet-NV was 77.3\%, improving by 7.4 points compared to the previous mathematical model-based method. Moreover, we have developed a product display robot using PYNet-NV. Our robot has successfully displayed 84.2\% of the products placed randomly. In the future, we plan to develop NNs that focus on surfaces that become newly visible after changing the viewpoint.

\section*{ACKNOWLEDGMENT}

A part of this work was supported by JSPS KAKENHI (grant number JP23K11157)


\begin{thebibliography}{99}

\bibitem{c1}R. B. Rusu, N. Blodow, and M. Beetz, “Fast point feature histograms (fpfh) for 3d registration,” in 2009 IEEE International Conference on Robotics and Automation, 2009, pp. 3212-3217. 
\bibitem{c2}L. S. F. Tombari, S. Salti, “Unique signatures of histograms for local surface description,” in European Conference on Computer Vision, 2010, pp. 356-369.
\bibitem{c3}F. Tombari, S. Salti, and L. Di Stefano, “A combined texture-shape descriptor for enhanced 3d feature matching,” in 2011 18th IEEE International Conference on Image Processing, 2011, pp. 809-812.
\bibitem{c4}G. A. Garcia Ricardez, S. Okada, N. Koganti, A. Yasuda, P. M. Uriguen, Eljuri, T. Sano, P.-C. Yang, L. El Hafi, M. Yamamoto, J. Takamatsu, and T. Ogasawara, “Restock and straightening system for retail automation using compliant and mobile manipulation,” Advanced Robotics, pp. 235-249, 2019.
\bibitem{c5}R. Sakai, S. Katsumata, T. Miki, T. Yano, W. Wei, Y. Okadome, N. Chihara, N. Kimura, Y. Nakai, I. Matsuo, and T. Shimizu, “A mobile dual-arm manipulation robot system for stocking and disposing of items in a convenience store by using universal vacuum grippers for grasping items,” Advanced Robotics, pp. 219-234, 2019.
\bibitem{pynet}K. Fujita, T. Tasaki, “PYNet: Poseclass and Yow Angle Output Network for Object Pose Estimation, ” Journal of Robotics and Mechatronics, Vol.35, No.1, pp.8-17, 2023.
\bibitem{dense_fusion}C. Wang, D. Xu, Y. Zhu, R. Martin-Martin, C. Lu, L. Fei-Fei, and S. Savarese, “DenseFusion: 6D object Pose Estimation by Iterative Dense Fusion,” International Conference on Computer Vision and Pattern Recognition, pp. 3343-3352, 2019.
\bibitem{ffb6d}Y. He, H. Huang, H. Fan, Q. Chen, J. Sun, “FFB6D: A Full Flow Bidirectional Fusion Network for 6D Pose Estimation,” International Conference on Computer Vision and Pattern Recognition, pp. 3002-3012, 2021.
\bibitem{long_edge}S. Kriegel, C. Rink, T. Bodenmuller, M. Suppa: “Efficient next-best-scan planning for autonomous 3D surface reconstruction of unknown objects”, J Real-Time Image Proc, Vol 10 ,No. 4, pp611-631, 2015.
\bibitem{outline}J. Hu, P. R.Pagila and Texas A\&M University , View Planning for Object Pose Estimation Using Point Clouds: An Active Robot Perception Approach, IEEE Robotics  and Automation Letters, Vol 7, No. 4, pp9248-9255, 2022
\bibitem{linemod}S. Hinterstoisser, V. Lepetit, S. Ilic, S. Holzer, G. Bradski, K. Konolige, and N. Navab, “Model based training, detection and pose estimation of texture-less 3D objects in heavily cluttered scenes,” Asian Conference on Computer Vision, pp. 548-562, 2012.
\bibitem{deep_im}Y. Li, G. Wang, X. Ji, Y. Xiang, and D. Fo, “DeepIM: Deep Iterative Matching for 6D Pose Estimation”,Vol 11210, pp695-711, 2018
\bibitem{rembg}X. Qin, Z. Zhang, C. Huang, M. Dehghan, O. R. Zaiane and M. Jagersand University of Alberta, Canada, “U2-net:Going deeper with nested u-structure for salient object detection,” Pattern Recognition, Vol 106, pp107404, 2020.
\bibitem{sam}A. Kirillov, E. Mintun, N. Ravi, H. Mao, C. Rolland, L. Gustafson, T. Xiao, S. Whitehead, A. C. Berg, W. Lo, P. Dollar, and R. Girshick, : “Segment Anything”, In International Conference on Computer Vision, 2023.
\bibitem{wrs}H. Okada, T. Inamura, and K. Wada, “What competitions were conducted in the service categories of the world robot summit?,” Advanced Robotics, vol. 33, pp. 900-910, 2019.

\end{thebibliography}
\end{document}